\documentclass{article} 
\usepackage[preprint]{tmlr}

\usepackage{amsmath,amsfonts,bm}

\def\eqref#1{equation~\ref{#1}}

\def\1{\bm{1}}

\DeclareMathAlphabet{\mathsfit}{\encodingdefault}{\sfdefault}{m}{sl}
\SetMathAlphabet{\mathsfit}{bold}{\encodingdefault}{\sfdefault}{bx}{n}

\usepackage[T1]{fontenc}
\usepackage{charter}
\usepackage{hyperref}
\usepackage{url}

\usepackage[utf8]{inputenc}
\usepackage[T1]{fontenc}
\usepackage{booktabs}
\usepackage{tabularx,array}
\usepackage{amsfonts}
\usepackage{nicefrac}
\usepackage{microtype}
\usepackage{xcolor}
\usepackage{graphicx}
\usepackage{amsmath, amssymb, amsthm, mathtools}
\usepackage{bm}
\usepackage{enumitem}
\usepackage{algorithm}
\usepackage{algorithmic}
\usepackage{caption}
\usepackage{subcaption}
\usepackage{multirow}
\usepackage{tcolorbox}
\usepackage{placeins}
\usepackage{xcolor}
\usepackage{pifont}
\usepackage{makecell}

\newcommand{\cmark}{\textcolor{green!60!black}{\ding{51}}}
\newcommand{\xmark}{\textcolor{red}{\ding{55}}}
\hypersetup{colorlinks=true, linkcolor=blue!70!black, citecolor=green!50!black, urlcolor=blue!60!black}

\title{PowerSim: Differentiable Physics Simulation and Rendering with Power Diagrams}

\author{\vspace{-20pt}
\\Trong-Tung Nguyen\quad\quad Anand Bhattad \\
[5pt]
Johns Hopkins University \\
[10pt]
\normalfont \textbf{Project Website}: \href{https://power-sim.github.io}{\texttt{https://power-sim.github.io}}
}

\usepackage[autostyle=true]{csquotes}

\begin{document}

\maketitle
\vspace{-0.65cm}
\begin{center}
\centering
    \includegraphics[width=\linewidth, trim={0cm 2cm 0cm 2cm}, clip]{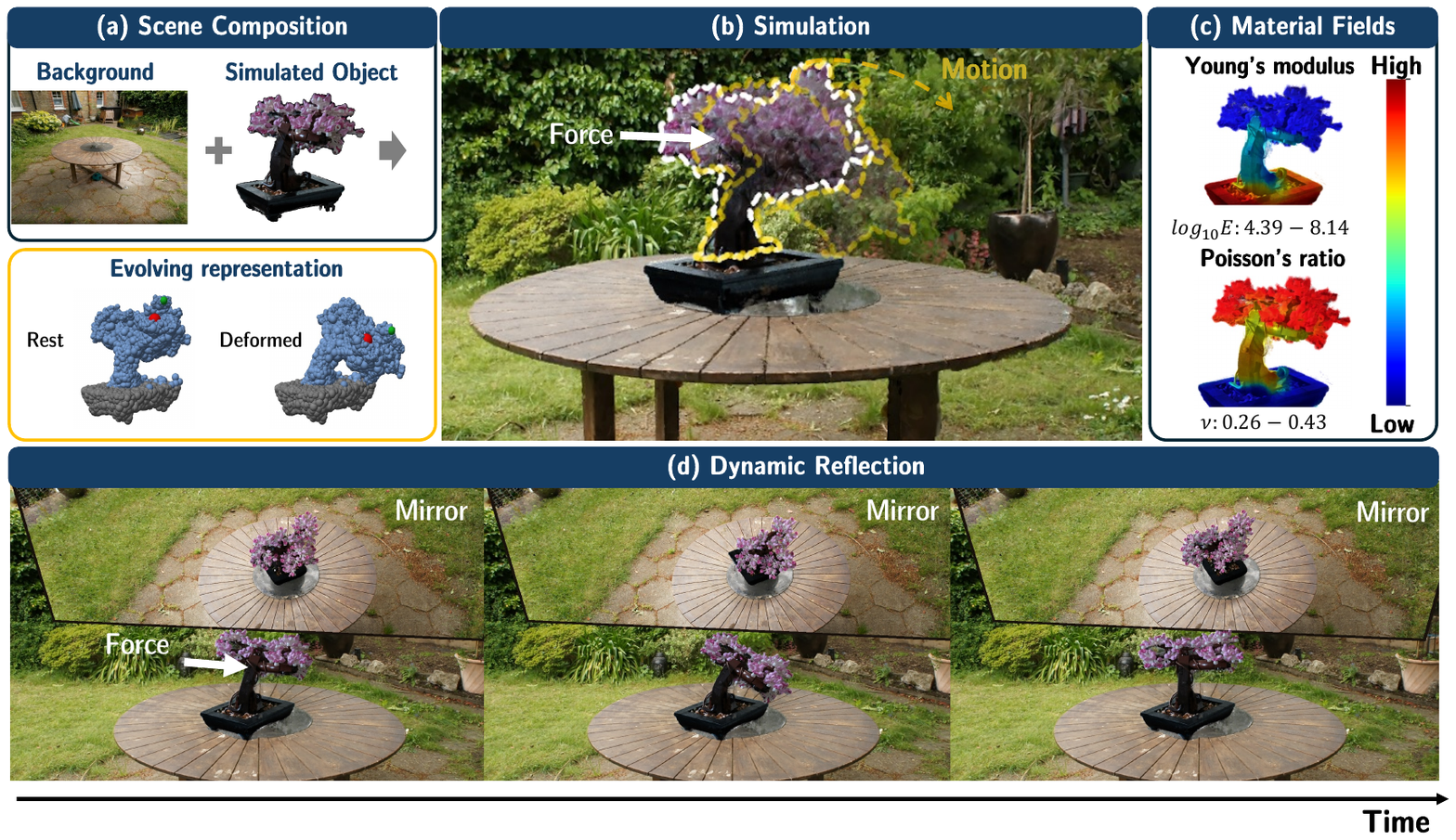}
    \vspace{-20pt}
 \captionof{figure}{\textbf{PowerSim unifies MPM simulation and rendering with bounded power diagrams in an end-to-end differentiable pipeline.}
(a) Objects from independent captures can be selected and composed into a scene.
(b) Primitive geometry and appearance evolve under applied forces.
(c) Spatially varying material properties can be estimated from video or assigned to control deformation.
(d) Ray-traced reflections follow the object's motion and deformation.}
    \label{fig:teaser}
    \vspace{-5pt}
\end{center}

\begin{abstract}
\looseness=-1 We introduce \textbf{PowerSim}, a method to bring physically grounded, differentiable dynamics to PowerFoam's power diagram based 3D representation. 
PowerSim directly couples a pre-trained PowerFoam scene to the Material Point Method (MPM) by exploiting a natural alignment between the two: the geometric and appearance properties of each primitive correspond closely to the quantities MPM already tracks as an object deforms. Consequently, simulated motion can drive the scene’s geometry and appearance directly, without an auxiliary representation in between. Built on this framework, we enable a range of applications on real and synthetic scenes: (1) simulating a static scene under user interaction, (2) recovering spatially varying material fields, (3) compositing primitives from independently captured scenes into a single simulation-ready scene and (4) ray-tracing reflections that update consistently as the object deforms. Our results suggest that PowerSim excels over previous frameworks for physically grounded dynamics, while unlocking unique advantages—such as secondary ray lighting effects on dynamic scenes. Results are best viewed on our project website: \url{https://power-sim.github.io/}
\end{abstract}
\vspace{-5pt}
\section{Introduction}
\begin{figure}[t!]
\vspace{-10pt}
    \centering
    \includegraphics[width=\linewidth, trim={1.9cm 2cm 1.3cm 3cm}, clip]{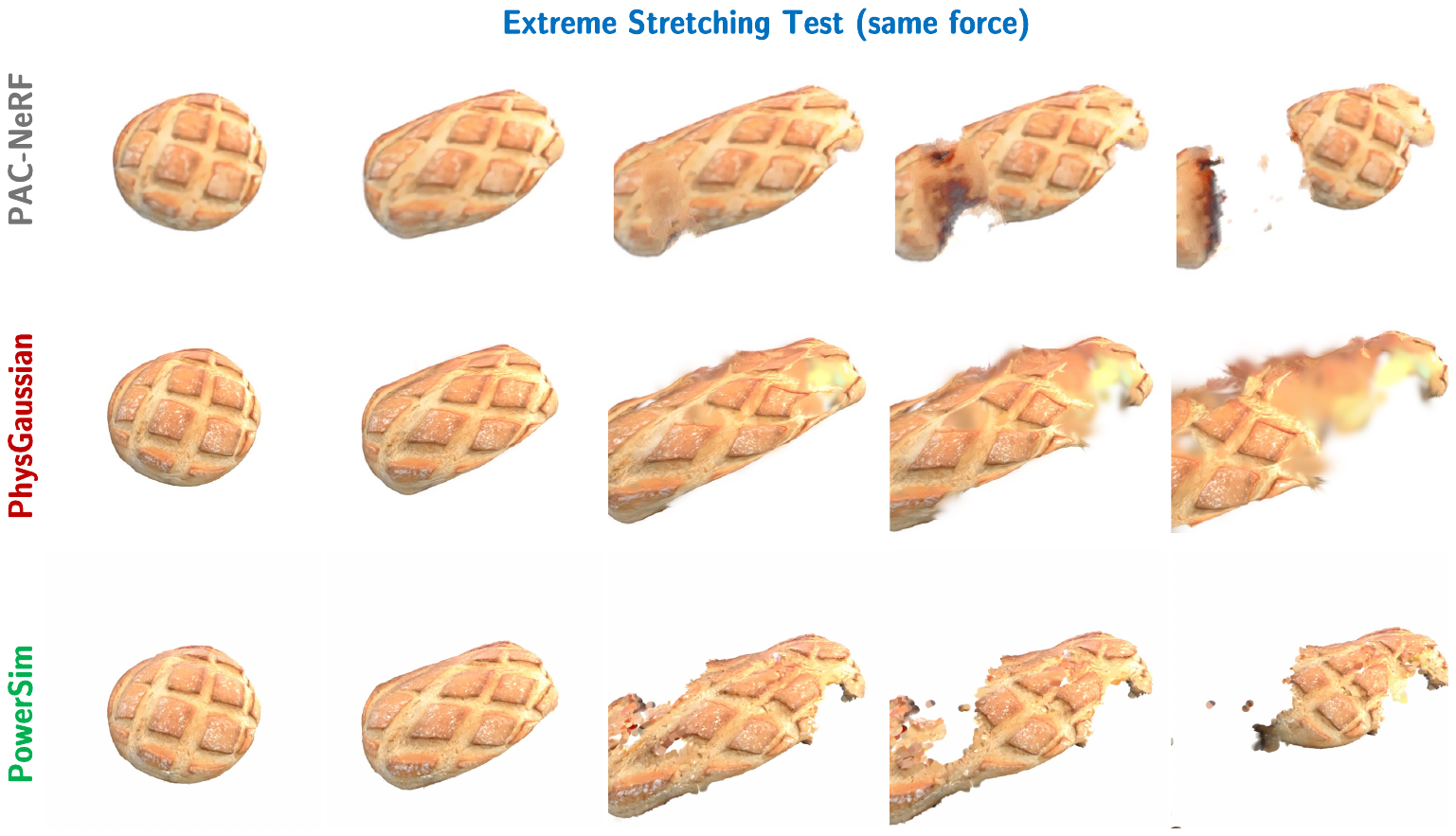}
\vspace{-12pt}
\caption{PowerSim partitions the reconstructed object volume into explicit cells, while PhysGaussian~\citep{xie2023physgaussian} uses Gaussian kernels that tend to concentrate near surfaces and optionally adds interior particles.
Under extreme stretching with the same applied force, PAC-NeRF loses texture and develops abrupt cut-like boundaries, while PhysGaussian exhibits pronounced blurring in stretched regions. PowerSim retains more surface detail and sharper boundaries during separation. 
}
    \label{fig:compare_with_physgauss}
\vspace{-12pt}
\end{figure}

Radiance field methods have made  3D scene capture routine~\citep{mildenhall2021nerf, kerbl20233d}. A handful of images are enough to render it from novel viewpoints. But without temporal data, they only give us one instant in time, and many interesting questions about a scene require us to infer what might happen next.
For example, what would happen if we replaced the vase on the table with a plant and pushed it? How would its reflection in a mirror change (Fig.~\ref{fig:teaser})? What would happen if we pulled a loaf of bread apart (Fig.~\ref{fig:compare_with_physgauss})? To answer these questions, we need a simulation that preserves the visual fidelity of the captured scene. To infer physical parameters from observed motion, we also need to make this simulation-and-rendering pipeline differentiable.

However, simulation and rendering place different demands on a scene representation. Continuum simulation requires material volumes and a treatment of contact and separation, while rendering requires geometry and appearance sufficient to generate images. 
Mesh-based pipelines start from explicit geometry for simulation and attach an appearance model to it \citep{mai2026radiance, held2025meshsplatting}. Gaussian-based pipelines start from a representation designed for rendering and adapt it for simulation using the Material Point Method (MPM)~\citep{xie2023physgaussian}.

But Gaussian mixtures do not explicitly specify non-intersecting volumes of materials, an explicit boundary surface or adjacency of primitives. PhysGaussian provides optional regularization of Gaussian anisotropy and filling of interior regions of objects with additional particles, which, however, do not lead to establishing explicit boundaries between primitives. Thus, specifying material volumes and masses, contacts, and refractions implies making some extra assumptions or using additional constructions beyond the Gaussian representation itself~\citep{moenne20243d}. 


PowerFoam~\citep{govindarajan2026powerfoam} reconstructs a scene as bounded power-diagram cells with explicit boundaries and neighbors, providing geometry for both simulation and rendering. But deforming a partition is harder than moving a set of points. Moving one site changes neighboring cell boundaries, and large deformations can change cell adjacency. The challenge is to update the primitives and their neighborhoods together as the object deforms.

To address this, we introduce \textbf{PowerSim}, which treats each PowerFoam primitive as an MPM material point. MPM updates its position, while the deformation gradient supplies a rotation for its dipole plane and appearance directions, and an isotropic radius scale that matches the local volume change. We rebuild adjacency as the primitives move and change size, keeping cell boundaries consistent with the updated representation. The result is a differentiable pipeline from forces to pixels: we can simulate a captured scene under new forces and backpropagate through simulation and rendering to estimate spatially varying material properties from monocular deformation videos. PowerSim improves average dynamic reconstruction quality and Young's modulus estimation over the compared methods, and preserves more surface detail under large stretching and twisting deformations (Fig.~\ref{fig:compare_with_physgauss}). We also demonstrate elastic motion, compression, granular collapse, and multi-object interactions with reflections that follow the simulated motion.

In summary, our contributions are:
\begin{enumerate}[leftmargin=1.8em, itemsep=0pt]
\item PowerSim, a differentiable framework coupling PowerFoam with MPM through updates to primitive geometry, appearance, and adjacency, supporting large deformation and separation.
\item Estimation of spatially varying material fields from monocular deformation videos through differentiable simulation and rendering.
\item Optimization-free selection of object primitives from multi-view masks, enabling scene composition and multi-object simulation with ray-traced reflections.
\end{enumerate}
\vspace{-5pt}

\section{Related Work}
\label{sec:related_works}

\paragraph{Physics-based scene simulation.}
\looseness=-1 A growing line of work couples reconstructed scenes to physics
solvers. PhysGaussian~\citep{xie2023physgaussian} treats
3D Gaussians as MPM particles, using the same primitives
for simulation and rendering.
PhysDreamer~\citep{zhang2024physdreamer},
DreamPhysics~\citep{huang2024dreamphysics}, and
Physics3D~\citep{liu2024physics3d} estimate material properties
using generated videos or video diffusion priors.
Spring-Gaus~\citep{zhong2024springgaus} instead fits a
spring-mass model to observed videos. PhysTwin \citep{jiang2025phystwin} reconstructs deformable objects and estimates dense physical properties from sparse interaction videos using spring-mass physics and Gaussian rendering. Meanwhile, GIC \citep{cai2024gic} estimates physical properties using a Gaussian-informed continuum, with surface and silhouette supervision, and NeuMA \citep{Cao_2024_NeuMA} learns corrections to material models and uses Particle-GS to propagate image gradients into the simulator. Other representations include PAC-NeRF's voxel--particle
coupling~\citep{li2023pac} and PhysConvex's deformable convex
primitives with reduced-order
simulation~\citep{wang2026physconvexphysicsinformed3ddynamic}. Pixie~\citep{le2025pixie} predicts material fields from visual features, while Vid2Sim~\citep{chen2025vid2sim} combines
feed-forward reconstruction with optimization.
PowerSim couples bounded power-diagram cells to MPM,
updating their geometry, appearance, and adjacency during
deformation.

\vspace{-5pt}
\paragraph{3D primitives selection.} Selecting the subset of representation primitives that belong to an object is a prerequisite task for editing or simulating reconstructed scenes. A common strategy attaches a learnable semantic attribute to each primitive and optimizes it through differentiable rendering against 2D supervision from foundation models such as SAM \citep{kirillov2023segany}. Gaussian Grouping \citep{ye2024gaussian} augments each Gaussian with an identity encoding that is rendered and classified by a linear layer supervised with view-consistent SAM masks. SAGA \citep{cen2025segment} distills SAM features into per-Gaussian affinity features for prompt-based selection, and LabelGS \citep{zhang2025labelgs} lifts multi-class pixel labels to individual Gaussians. SemanticFoam \citep{semanticfoam2026} addresses this problem on a different representation RadiantFoam  \citep{govindarajan2025radiant} by learning per-cell
identity encodings and regularizing them with a total-variation loss over the Voronoi adjacency graph \citep{govindarajan2025radiant}. Beyond methods that learn optimized semantic fields,
FlashSplat \citep{flashsplat} derives Gaussian labels using a closed-form solution, while masked-gradient voting \citep{joji2024gradient} and gradient-weighted back-projection \citep{joseph2024gradientweightedfeaturebackprojection} transfer 2D masks and features without training per scene. In contrast, we propose render-weighted voting for PowerFoam cells, label propagation
for unseen primitives by adjacency between cells in a training-free manner rather than optimization-based like SemanticFoam \citep{semanticfoam2026}, and perform simulation using the selected primitives for scene composition.

\vspace{-5pt}
\paragraph{Rendering and scene representations.}
3DGRT and 3DGUT support ray tracing and secondary-ray effects
with Gaussian representations~\citep{moenne20243d,wu20253dgut}.
PowerFoam~\citep{govindarajan2026powerfoam} supports both
rasterization and ray tracing using bounded power-diagram cells.
We build on this representation to simulate captured objects,
estimate their material properties, and render reflections
that follow their deformation.

\section{Preliminaries}
\label{sec:background}
\textbf{PowerFoam}~\citep{govindarajan2026powerfoam} represents a scene
as bounded power-diagram cells. Each primitive has a center
$\mathbf{p}_i \in \mathbb{R}^3$ and a scalar radius
$\mathbf{r}_i$, which define its power cell:
\vspace{-5pt}
\begin{equation}
\mathbf{P}_i =
\left\{\mathbf{x} \in \mathbb{R}^3 \;\middle|\;
\|\mathbf{x}-\mathbf{p}_i\|^2-\mathbf{r}_i^2
\leq
\|\mathbf{x}-\mathbf{p}_j\|^2-\mathbf{r}_j^2,\ \forall j
\right\}.
\label{eq:def_powerfoam}
\end{equation}
The cell is bounded by intersecting it with the ball
$\mathbf{B}_i=\mathbf{B}(\mathbf{p}_i,\mathbf{r}_i)$.
These cells provide explicit boundaries and neighbor relations.

\textit{Geometry.}
Each primitive contains a dipole plane through $\mathbf{p}_i$,
with a material side of density $\boldsymbol{\sigma}_i$
and an empty side.
A quaternion $\mathbf{q}_i$ defines the plane's orthonormal
frame: normal $\mathbf{n}_i$ and tangents
$\mathbf{t}_i,\mathbf{b}_i$.
To represent surface detail, each plane carries $K$ detail
sites with local in-plane coordinates
$\mathbf{s}_{i,k}\in\mathbb{R}^2$ and normal displacements
$d_{i,k}\in\mathbb{R}$. Their world positions are
\vspace{-5pt}
\begin{equation}
\mathbf{x}_{i,k}
=
\mathbf{p}_i
+\mathbf{r}_i
\big(
\mathbf{s}_{i,k}^{1}\mathbf{t}_i
+\mathbf{s}_{i,k}^{2}\mathbf{b}_i
\big)
+d_{i,k}\mathbf{n}_i.
\label{eq:pin_world_pos}
\end{equation}
Interpolating the displacements gives a height field over
the plane.

\textit{Appearance.}
Each detail site stores $L$ world-space appearance directions
$\boldsymbol{\alpha}_{i,k}^{(l)}$ and corresponding RGB colors
$\mathbf{c}_{i,k}^{(l)}$.
The viewing direction determines how these colors are blended;
a second interpolation blends colors across detail sites.
We give both interpolation rules in
Appendix~\ref{sec:powerfoam_details}.

We write the reconstructed rest scene as $\mathcal{S}_0$,
with image $I=\mathcal{R}(\mathcal{S}_0)$.
For each primitive, $\mathbf{s}_i$, $d_i$, and
$\boldsymbol{\alpha}_i$ collect its detail-site coordinates,
displacements, and appearance directions.
During simulation, we update
$\{\mathbf{p}_i^t,\mathbf{r}_i^t,\mathbf{q}_i^t,
\boldsymbol{\alpha}_i^t\}$ and keep the remaining attributes
fixed. The local detail sites follow the primitive through
Eq.~\ref{eq:pin_world_pos}, while the world-space appearance
directions must be rotated explicitly.

\vspace{2pt}
\textbf{Material Point Method} (MPM)~\citep{jiang2016material}
simulates a continuum body using particles and a background grid. Each particle carries position $\mathbf{x}_p$, velocity
$\mathbf{v}_p$, mass $m_p$, deformation gradient $\mathbf{F}_p$,
and an affine velocity matrix $\mathbf{C}_p$.
Its material parameters are
$\boldsymbol{\theta}_p=(E_p,\nu_p)$, where $E_p$ is Young's
modulus and $\nu_p$ is Poisson's ratio.
The deformation gradient tracks local rotation, stretch,
and shear and determines stress through the material's
constitutive model. Each step transfers particle mass and momentum to the grid,
updates grid velocities under internal and external forces,
and transfers the result back to the particles.
Writing the simulator as $\mathcal{M}_{\boldsymbol{\theta}}$,
one substep gives
\begin{equation}
\big(
\mathbf{x}^{t+1},\mathbf{v}^{t+1},
\mathbf{F}^{t+1},\mathbf{C}^{t+1}
\big)
=
\mathcal{M}_{\boldsymbol{\theta}}
\big(
\mathbf{x}^{t},\mathbf{v}^{t},
\mathbf{F}^{t},\mathbf{C}^{t}
\big).
\label{eq:mpm_step}
\end{equation}
The full simulation loop is given in
Appendix~\ref{sec:appendix}.

\section{Method}
\begin{figure}[t]
    \centering
    \includegraphics[width=\linewidth, trim={0.6cm 10.5cm 0cm 4cm}, clip]{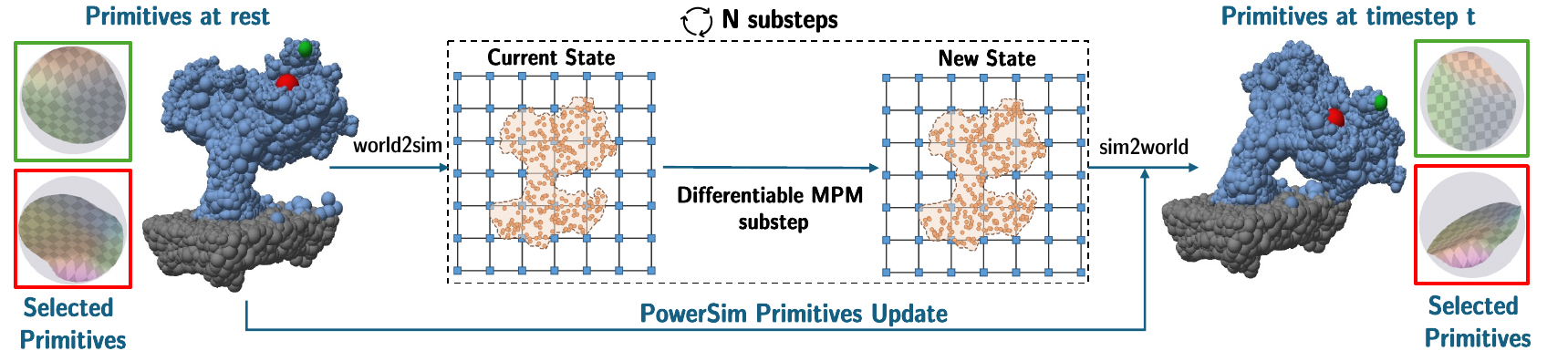}
    \vspace{-10pt}
    \caption{\textbf{PowerSim overview.} After $N$ MPM substeps, we update primitive positions, isotropically scale their radii to match local volume change, and rotate their dipole frames and appearance directions. We then rebuild adjacency for the updated primitives. Insets show two selected primitives before and after deformation.}
    \label{fig:evolve_sites}
    \vspace{-5pt}
\end{figure}

Given a pretrained PowerFoam scene, we update its primitives using MPM simulation output (\S\ref{sec:evolve_sites}; Fig.~\ref{fig:evolve_sites}). We differentiate through simulation, primitive updates, and rendering to estimate spatially varying material properties from monocular deformation videos (\S\ref{sec:material_estimation}). We also describe an optimization-free method to select simulation-ready object primitives from multi-view masks (\S\ref{sec:foam_selection}).

\vspace{-5pt}
\subsection{PowerSim: Physics-grounded PowerFoam Simulation with MPM}
\label{sec:evolve_sites}
A pretrained PowerFoam scene is purely static: every primitive carries geometry and appearance but no physical quantities, since nothing in photometric reconstruction needs them. In this work, we turn these primitives into continuum dynamic bodies by treating each primitive as an MPM particle and augmenting each with exactly the physical state MPM requires. Each augmented primitive thus carries geometry,
appearance, and physical properties at once, and as it moves, all three must remain consistent with the physics-grounded simulation. The physical state $\mathbf{X}_t = \{\mathbf{x}_i^t, \mathbf{v}_i^t, \mathbf{F}_i^t, \mathbf{C}_i^t\}$ is advanced by MPM itself as discussed in Eq.\ref{eq:mpm_step}; what remains is to propagate these physical state update to the primitive channels. As shown in Fig.\ref{fig:evolve_sites}, we derive an update operator $\mathcal{U}$ that, given MPM's simulation output at time $t$, maps a primitive's rest-state channels to their deformed versions:
\begin{equation}
\mathcal{U}:\ \big(\mathbf{p}_i^0,\ \mathbf{r}_i^0,\ \mathbf{q}_i^0,\ \boldsymbol{\alpha_i^0}\big)
\ \longmapsto\
\big(\mathbf{p}_i^t,\ \mathbf{r}_i^t,\ \mathbf{q}_i^t,\ \boldsymbol{\alpha}_i^t\big),
\label{eq:update_operator}
\end{equation}
\paragraph{Volume and Mass.} We partition the MPM background grid into voxels of side length $\Delta x$ and divide each occupied voxel's volume $\Delta x^3$ equally among the particles it contains to obtain the rest volume $V_i^0$. Particle's mass is then $m_i = \rho V_i^0$ for a uniform density
$\rho$ per scene. $V_i^0$ is computed once at rest, and subsequent volume change is carried by
$J = \det(\mathbf{F})$.
\paragraph{Primal Sites Update.} We let primitives' centers evolve by transforming their rest position $\mathbf{p}_i^0$ to MPM's own coordinate frame, $\mathbf{x}_i^0 = \Phi(\mathbf{p}_i^0)$; from there, MPM's loop runs through several $N$ substeps to evolve $\mathbf{x}_i^t$ directly, and we recover the primitive's world-space position at any time via the inverse map $\mathbf{p}_i^t = \Phi^{-1}(\mathbf{x}_i^t)$. Every other primitive property updates differently, by leveraging the local deformation information MPM tracks throughout the simulation: the \textbf{deformation gradient} $\mathbf{F}_i^t$, representing the primitive's accumulated deformation relative to its rest state. Physically, $\mathbf{F}_i^t$ encodes a mix of rotation, stretch, and shear accumulated in the material around primitive $i$. We first apply polar decomposition to $\mathbf{F}_i^t$ to obtain its rotation and stretch components: 
\begin{equation} 
\mathbf{F}_i^t = \mathbf{R}_i^t\, \mathbf{S}_i^t, \qquad \mathbf{F}_i^t = \mathbf{U}_i \mathbf{\Sigma}_i \mathbf{V}_i^\top,\ \ \mathbf{R}_i^t = \mathbf{U}_i \mathbf{V}_i^\top, 
\label{eq:polar} 
\end{equation} 
Here $\mathbf{R}_i^t$ is a rigid rotation and $\mathbf{S}_i^t$ a symmetric stretch, with singular values $\mathbf{\Sigma}_i = \mathrm{diag}(\sigma_i^{(1)}, \sigma_i^{(2)}, \sigma_i^{(3)})$. Next, we map $\mathbf{R}_i^t$ and $\mathbf{S}_i^t$ onto PowerFoam's remaining primitive channels.

\paragraph{Power Radii Update.} PowerFoam's scalar radius has no anisotropic degree of freedom, so we project
$\mathbf{S}_i^t$ onto the isotropic scale that reproduces the same local volume change:
\begin{equation}
\bar\sigma_i^t = \big(\sigma_i^{(1)}\sigma_i^{(2)}\sigma_i^{(3)}\big)^{1/3},
\qquad
\mathbf{r}_i^t = \mathbf{r}_i^{(0)}\cdot \bar\sigma_i^t.
\label{eq:radius_update}
\end{equation}
In this way, we deliberately leave each individual primitive to stay isotropic and compact, anisotropy arises from the developing geometry of the power diagram.
On the other hand, the use of anisotropic 3DGS may lead to absorbing deformation through stretching primitives, thus resulting in very elongated primitives and artifacts in the case of significant deformation as shown in Fig.\ref{fig:compare_with_physgauss}. PowerSim, however, offloads the problem of anisotropic shape changes to the geometry of neighboring cells, thus making the representation possible to be significantly deformed without elongating the primitives.

\paragraph{Dipole Planes Update.} Dipole plane's normal and local reference frame $(\mathbf{n}_i, \mathbf{t}_i, \mathbf{b}_i)$ are parameterized by a single quaternion $\mathbf{q}_i^0$. We update $\mathbf{q}_i^0$ correspondingly with the rotation $\mathbf{R}_i^t$ obtained from the polar decomposition, since this naturally rotates the primitive's geometry to the corresponding angle without disturbing the relative orthogonality of $\mathbf{n}_i, \mathbf{t}_i, \mathbf{b}_i$. The quaternion update is achieved via 
\begin{equation}
\mathbf{q}_i^t = \mathbf{q}_i^0 \otimes \Delta \mathbf{q}_i^t, \qquad \Delta \mathbf{q}_i^t = \mathrm{quat}\!\big(\mathbf{R}_i^{t\top}\big),
\label{eq:quat_update}
\end{equation}
composed with the rest orientation $\mathbf{q}_i^0$ as the left operand. Note that the normal and tangents are encoded as row vectors of the rotation matrix, which is what leads to the order of multiplication and to composing with the \emph{inverse} (transpose) of $\mathbf{R}_i^t$.

\paragraph{Detail Sites Update.} From \S \ref{sec:background}, each dipole plane carries $K$ detail sites with local coordinates $\mathbf{s}_i$, displacement $d_i$, and radiance axis $\boldsymbol{\alpha}_i$. The first two are expressed relative to the primal site's position, radius, and frame, so they need no update: re-evaluating Eq.~\ref{eq:pin_world_pos} against the deformed primal site places them correctly for free. The radiance axis is a world-space direction and inherits nothing from the frame, so we update it explicitly, rotating the axis
while holding its color coefficients fixed. We require that a co-rotating viewer sees no change: for every world direction $\mathbf{d}$, the shading response must match what the rest-pose primitive produces for the back-rotated direction $\mathbf{R}_i^{t\top}\mathbf{d}$,
\begin{equation}
\big\lVert \mathbf{d} - \boldsymbol{\alpha}_i^t \big\rVert_2 = \big\lVert \mathbf{R}_i^{t\top} \mathbf{d} - \boldsymbol{\alpha}_i^0 \big\rVert_2,
\end{equation}
which, by orthogonality of $\mathbf{R}_i^t$, gives
\vspace{-5pt}
\begin{equation}
\boldsymbol{\alpha}_i^t = \mathbf{R}_i^t \boldsymbol{\alpha}_i^0.
\label{eq:axis_update}
\end{equation}

\paragraph{Topology Update.} Since primitives change every step, neighbor relations are created and broken as the object deforms, and a stale topology would
clip cells against the wrong neighbors. We therefore rebuild the adjacency after every update as the \v{C}ech complex of the primitives' bounding spheres, which is the same structure PowerFoam already uses for rendering.

\subsection{Material Field Optimization }
\label{sec:material_estimation}
\begin{figure}
\vspace{-15pt}
    \centering
    \includegraphics[width=1.02\linewidth, trim={0.6cm 4.3cm 0cm 2cm}, clip]{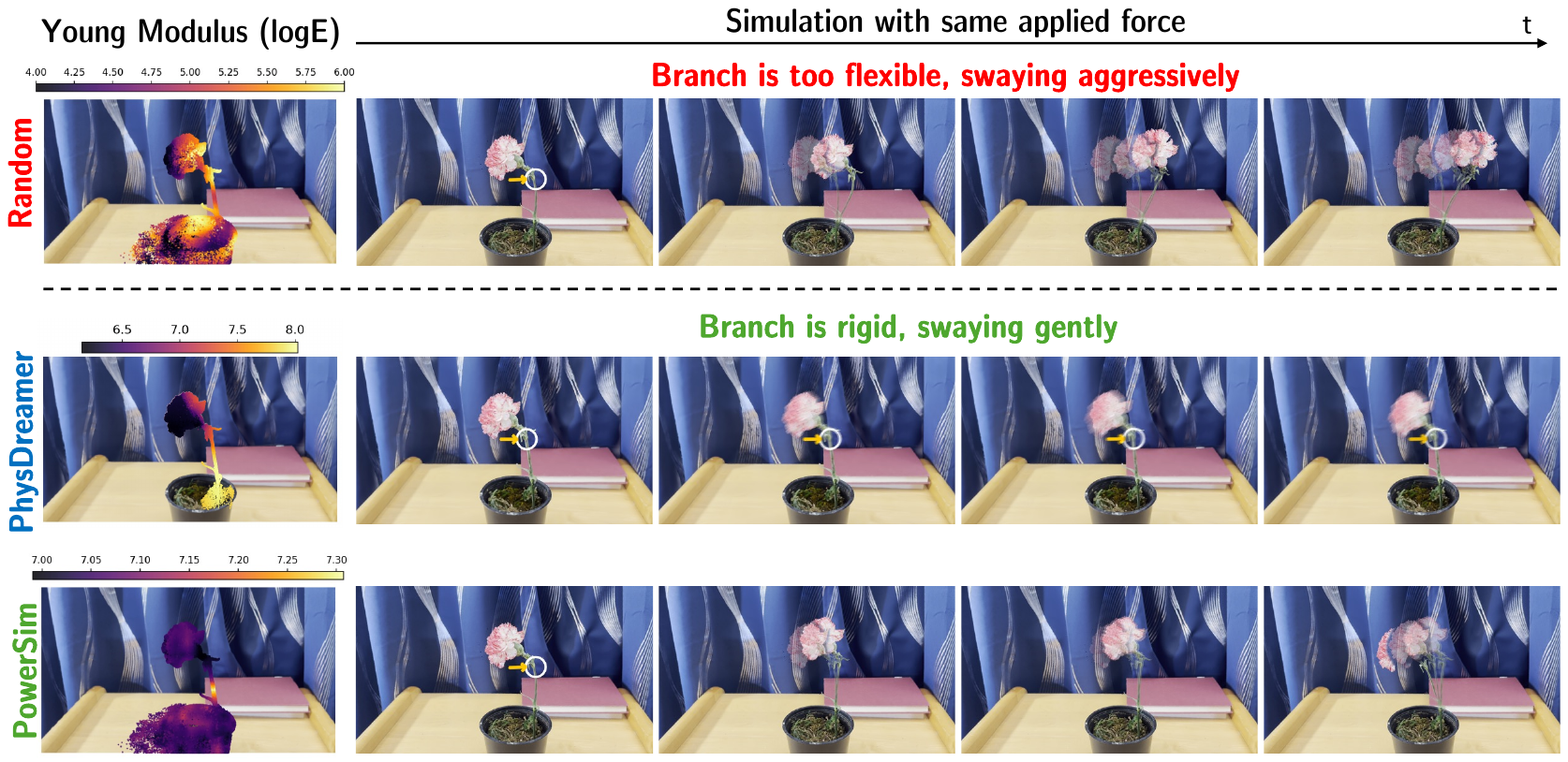}
\vspace{-15pt}
\caption{\textbf{Material estimation changes the simulated response.}
Under the same applied force, random material assignment produces large stem bending, while PhysDreamer produces little bending. PowerSim produces back-and-forth motion with visible stem deformation.
The left column shows Young's modulus; the remaining columns show matching simulation times.}
    \label{fig:material_estimate}
\vspace{-12pt}
\end{figure}
Coupled with MPM, PowerFoam primitives evolve under physically grounded rules; what remains is to determine the material that drives them. Their effect is not subtle: Fig. \ref{fig:material_estimate} contrasts the unrealistic motion produced by a wrong assignment (top row) with the plausible deformation obtained from correct parameters (third row). Manually assigning them, however, demands expert knowledge, does not scale beyond a handful of hand-tuned demo scenes, and is simply unavailable for objects reconstructed from video alone. We instead recover them directly via optimization from an observed video of the object deforming, after which new forces can be applied for new simulation.

\paragraph{Problem setup.} \looseness=-1 Given a reference video $\{I_t\}_{t=1}^T$ of an object in motion, which is captured directly or synthesized by a video generation model, our goal is to estimate a spatial material field, $\boldsymbol\theta_i = \{\theta_i\}_{i=1}^N$, where $\boldsymbol\theta_i = (E_i, \nu_i)$ indicates the Young's modulus and Poisson's ratio, such that PowerSim's simulated results reproduce the observed motion. Because the impulse setting the object in motion is generally unobserved, we optimize each primitive's initial velocity $\mathbf{v}_i^0$ alongside the material. 

\paragraph{Optimization.} At each training iteration, we run the full forward pipeline -- MPM simulation, our primitive-channel update, and rendering -- to obtain a sequence of simulated frames. Here we abuse $t$ to index video frames rather than substeps: advancing from frame $t$ to
$t{+}1$ takes $N_{\mathrm{sub}} = \Delta t_{\mathrm{frame}} / \Delta t_{\mathrm{sub}}$ MPM
substeps, so one simulated frame of the pipeline is
\begin{equation}
\hat{I}_{t+1} = \big(\mathcal{R} \circ \mathcal{U} \circ \mathcal{M}_{\boldsymbol{\theta}}^{N_{\mathrm{sub}}}\big)(\mathbf{X}_t),
\label{eq:rollout}
\end{equation}
where $\mathcal{M}_{\boldsymbol{\theta}}^{N_{\mathrm{sub}}}$ advances the physical state to
$\mathbf{X}_{t+1}$, $\mathcal{U}$ maps it -- with the rest state $\mathcal{S}_0$ held fixed -- to
the primitive state $\mathcal{S}_{t+1} = \mathcal{U}(\mathcal{S}_0, \mathbf{X}_{t+1})$, and
$\mathcal{R}$ renders it. The rollout starts from rest,
$\mathbf{X}_0 = \big(\Phi(\mathbf{p}^0),\ \mathbf{v}^0,\ \mathbf{I},\ \mathbf{0}\big)$. Because $\mathcal{R}$, $\mathcal{U}$, and $\mathcal{M}$ are each differentiable, so is Eq.~\ref{eq:rollout}. To update neighbors, we follow similarly to PowerFoam's training and rebuild adjacency at every frame from detached copies of the deformed sites. Then the rendering uses the original gradient-carrying sites with the updated adjacency list, neighborhood relations determine cell clipping without contributing gradients. This allows photometric losses on rendered frames to backpropagate to the unknowns. We compare each simulated frame against its reference video under:
\begin{equation}
    \mathcal{L}(\hat{I}_t, I_t) = (1-\lambda)\,\lVert \hat{I}_t - I_t \rVert_2^2 + \lambda\,\big(1 - \mathrm{SSIM}(\hat{I}_t, I_t)\big),
\label{eq:material_loss}
\end{equation}
so that $\partial\mathcal{L}/\partial\boldsymbol{\theta}$ and
$\partial\mathcal{L}/\partial\mathbf{v}^0$ pass from the rendered pixels through $\mathcal{R}$ and
$\mathcal{U}$ into the MPM rollout, and the unknowns are updated by the gradient descent. Following PhysDreamer \citep{zhang2024physdreamer}, we apply truncated backpropagation-through-time (BPTT) to avoid gradient explosion/vanishing. Rather than fitting $\boldsymbol{\theta}$ and $\mathbf{v}^0$ jointly, we optimize in two stages, recovering the initial velocity first and the material afterwards (more details in \S\ref{sec:appendix}). 

\subsection{Dynamic Primitives Selection via Render-weighted voting}
\label{sec:foam_selection}
A reconstructed real-world scene rarely contains a single object: a single kitchen scene contains thousands of primitives covering different objects, yet a simulation concerns one of them. This requires identifying the simulated primitives and assigning them material properties $\boldsymbol{\theta}_i$, leaving the rest as static geometry that is rendered but not simulated. However, curating these primitives by hand in 3D is tedious, and hence we tackle this issue with a simple method that leverages multi-view 2D segmentation masks at test-time, making it natural for picking primitives ready for simulation.

\paragraph{Render-weighted voting.} The user specifies the object through 2D segmentation masks
$\{M_j\}_{j\in\mathcal{J}}$ over a subset $\mathcal{J}$ of the training views, where $M_j(u)\in\{0,1\}$ marks whether pixel $u$ belongs to the object; off-the-shelf models can easily produce these from a text prompt or a click \citep{kirillov2023segany,liu2023grounding,cheng2023segment}. We seek a per-primitive label $\ell_i\in\{0,1\}$ whose selected primitives, rendered alone, reproduce $M_j$ on the annotated views and stay consistent on the rest. SemanticFoam \citep{semanticfoam2026} solves this problem by optimization and supervises the rendered feature against the masks; we instead read it off a trained checkpoint, since PowerFoam's rasterizer composites any per-primitive quantity the same way it
composites color. Replacing colors with scalar features $f_i$, the rendered feature at pixel $u$ of view $j$ is
\vspace{-5pt}
\begin{equation}
\hat{F}_j(u) = \sum_{i=1}^{N} w_{i,j}(u)\, f_i,
\label{eq:feature_render}
\end{equation}
where $w_{i,j}(u)$ is the compositing weight of primitive $i$ along the ray through $u$; which is the weight with which its color reaches the pixel. A
primitive's membership then follows from how much of its rendered contribution falls inside the masks versus outside, accumulated over views:
\begin{equation}
    s_i^{+} = \sum_{j\in\mathcal{J}}\sum_{u} w_{i,j}(u)\, M_j(u), \qquad
    s_i^{-} = \sum_{j\in\mathcal{J}}\sum_{u} w_{i,j}(u)\, \big(1 - M_j(u)\big),
    \label{eq:vote}
\vspace{-5pt}
\end{equation}

\begin{figure}[t!]
    \centering
    \includegraphics[width=\linewidth, trim={0.75cm 11cm 0.3cm 3.5cm}, clip]{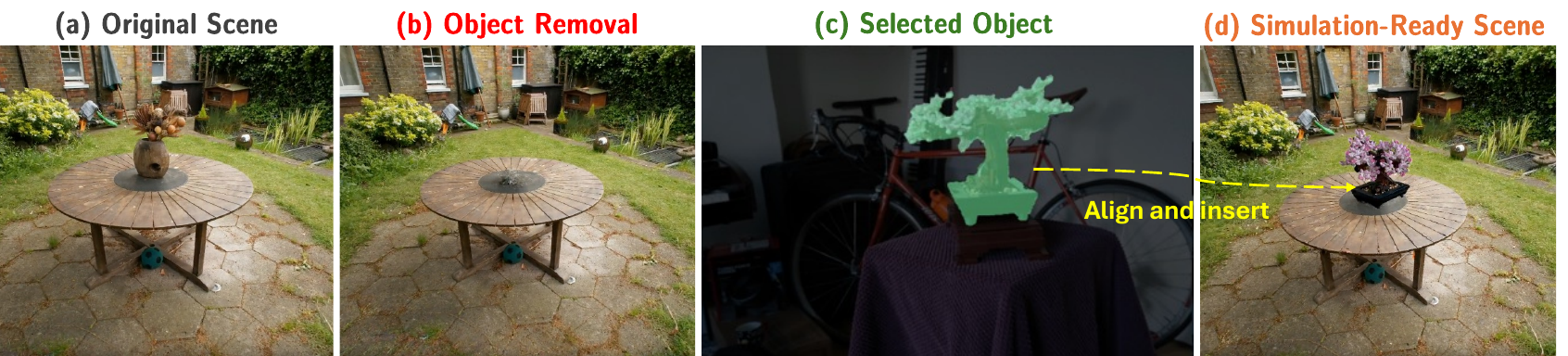}
    \vspace{-15pt}
\caption{\textbf{Scene composition.} Our optimization-free primitive selection enables removing the vase from a captured scene (a--b) and selecting a plant from another capture (c, green). We align and insert the selected primitives to create a scene ready for simulation (d).}
    \label{fig:foam_edit}
\vspace{-10pt}
\end{figure}
We select primitive $i$, setting $\ell_i = 1$, when $s_i^{+} > \beta\, s_i^{-}$, with $\beta < 1$
discounting the outside vote since masks tend to miss thin structures (We provide ablation of $\beta$ in \S\ref{subsec:additional_results}).

Computing $\{s_i^{+}, s_i^{-}\}$ never requires the weights themselves. Since Eq. \ref{eq:feature_render} is linear in $f$, the vote of primitive $i$ in view $j$ is a
derivative of the rendered feature image:
\begin{equation} 
    \sum_{u} w_{i,j}(u)\, M_j(u) = \frac{\partial}{\partial f_i}\sum_{u} \hat{F}_j(u)\, M_j(u). 
\label{eq:vote_grad} 
\end{equation} 
We therefore render the feature image, sum it over the masked pixels, and differentiate that scalar with respect to $f$: the gradient on $f_i$ is exactly primitive $i$'s vote, so one backward pass per view yields all $N$ votes at once, and $s_i^{-}$ follows with $1 - M_j$. Primitives no view observes ($w_{i,j}(u) = 0$ everywhere, hence $s_i^{+} = s_i^{-} = 0$) inherit the majority label of their power-diagram neighbors, using the adjacency the renderer already maintains. The procedure needs no training and no checkpoint change, and handles several objects by voting one score per class and taking the arg-max.

\vspace{-8pt}
\paragraph{From selection to simulation and editing.} The selected primitives become material points, receiving $\boldsymbol{\theta}_i$ --- a user-specified material or a field recovered by \S\ref{sec:material_estimation} --- while the rest stay fixed and are rendered alongside them. The selection also supports editing. Removal deletes the selected primitives and rebuilds the power-diagram adjacency as shown in Fig.\ref{fig:foam_edit}. Insertion brings cropped primitives into a target scene of unrelated frame and scale. The two are aligned by a simple transform: rotation, translation, and scaling, which is followed by adjacency rebuilt over the merged set. Replacement
is removal followed by insertion; Fig.\ref{fig:foam_edit} shows the vase replaced.

\section{Experimental Results}
\begin{figure}[t!]
    \centering
    \includegraphics[width=\linewidth, trim={2.1cm 0.1cm 1.3cm 0.1cm}, clip]{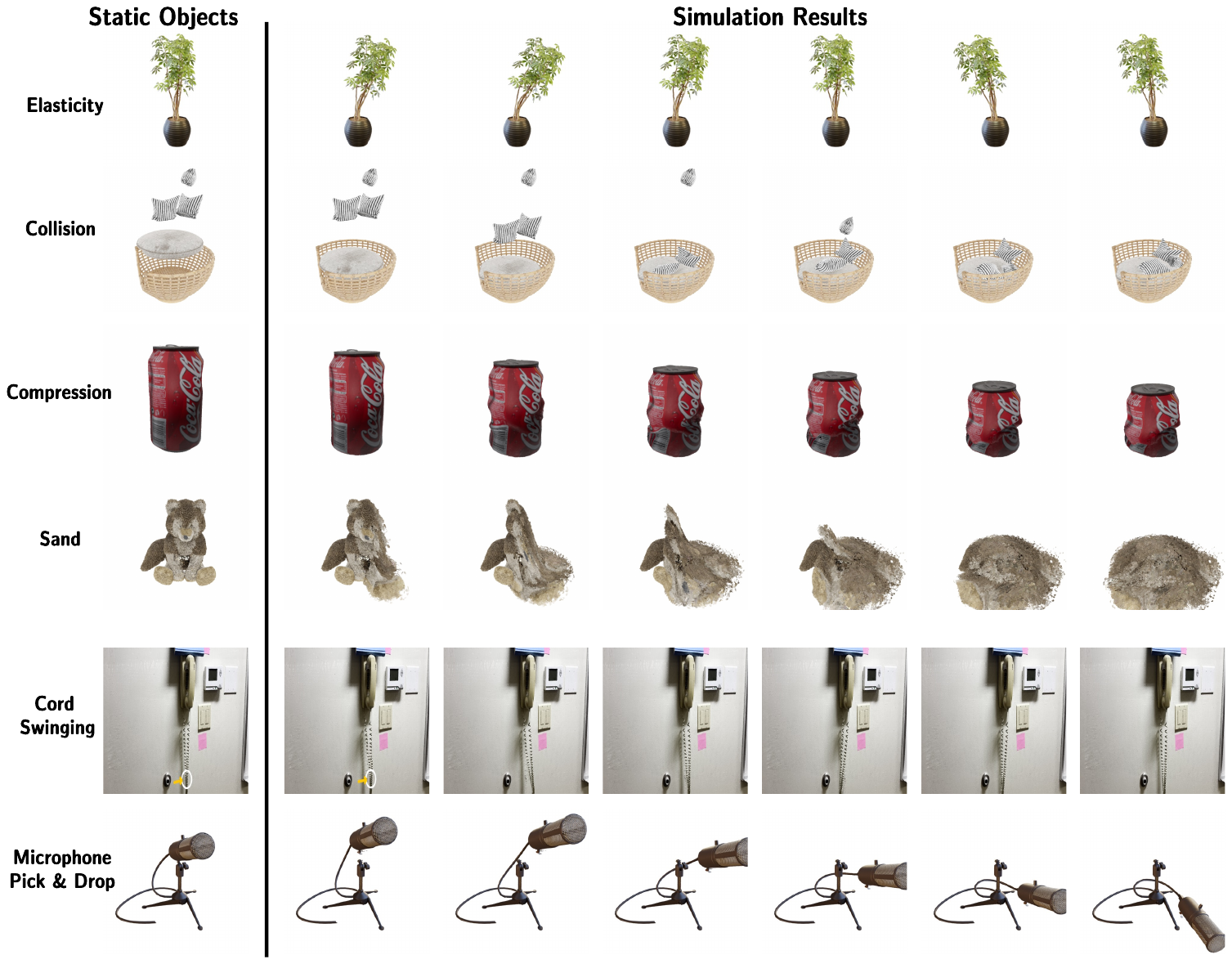}
\vspace{-25pt}
    \caption{\textbf{Diverse Material Behaviors.} Simulation results on different material settings.}
    \label{fig:more_sim}
\end{figure}

\begin{figure}[t!]
    \centering  \includegraphics[width=\linewidth, trim={0cm 5cm 0cm 5cm}, clip]{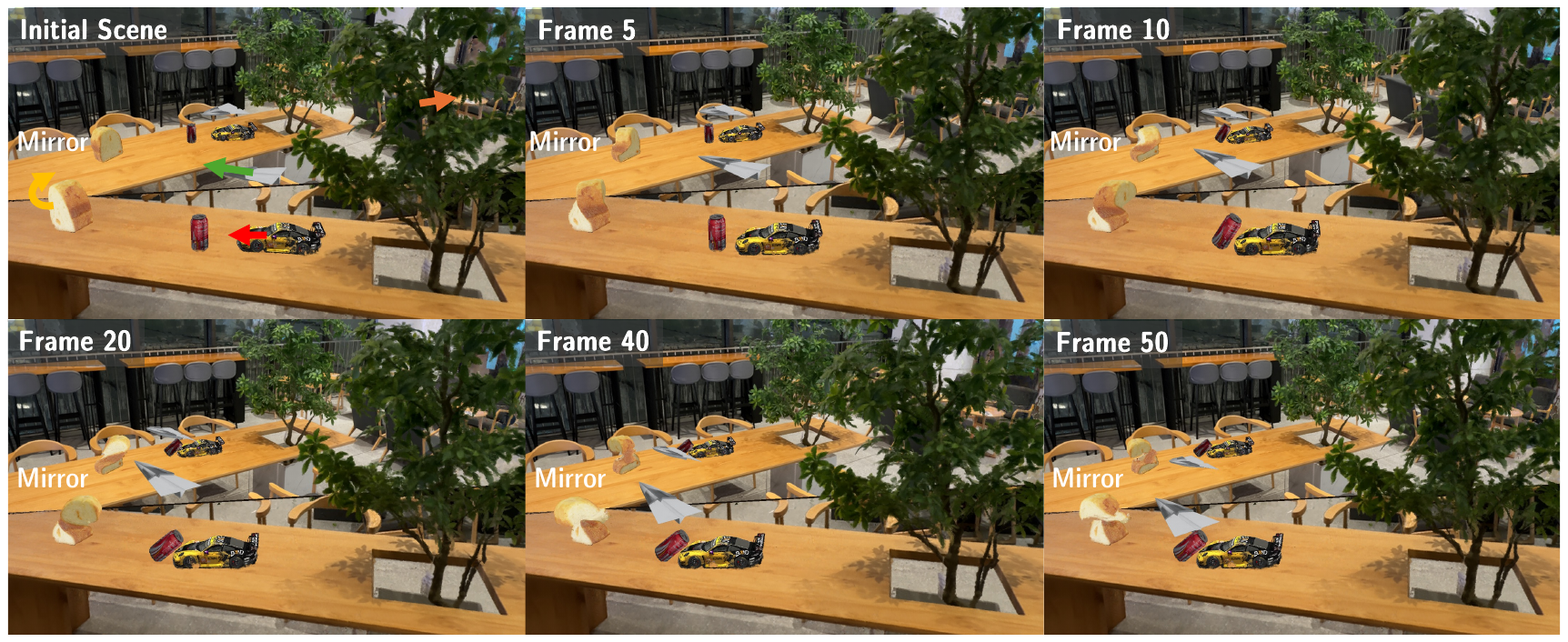}
\vspace{-15pt}
\caption{\textbf{Ray tracing on dynamic scenes.}
Independently captured objects are composited and simulated together, with mirror reflections that follow their motion and deformation.}
    \label{fig:multi_physics}
\vspace{-12pt}
\end{figure}

We evaluate simulation quality and material estimation against prior physics-coupled pipelines. We also demonstrate varied object dynamics (Fig.~\ref{fig:more_sim}) and secondary-ray effects in dynamic scenes (Fig.~\ref{fig:multi_physics}). Primitive selection is evaluated in \S\ref{subsec:additional_results}. We provide more simulation results in \S\ref{sec:appendix}.

\vspace{-8pt}
\paragraph{Benchmarks and Metrics.} To assess both physics simulation and material estimation, we use Vid2Sim benchmark \citep{chen2025vid2sim} which includes 12 objects from GSO (Google Scanned Object) \citep{downs2022google}, each dropped under gravity and
simulated with FEM with known ground-truth material parameters. We evaluate dynamic reconstruction quality against the reference videos via PSNR and SSIM \citep{wang2004image}, and material estimation by the absolute error of the recovered $\log_{10}E$ and $\nu$.

\paragraph{Baselines.} For physics simulation and material estimation, we compare against representative physics-coupled pipelines on other representations: PhysDreamer \citep{zhang2024physdreamer} on 3DGS, and PAC-NeRF \citep{li2023pac} on NeRF. PhysDreamer extends PhysGaussian \citep{xie2023physgaussian} with additional material optimization. For 3D segmentation, we compare against Semantic Foam \citep{semanticfoam2026}, which requires optimization to obtain a per-primitive semantic field on the same foam family, and against the Gaussian-based Gaussian Grouping \citep{ye2024gaussian}, and LabelGS \citep{zhang2025labelgs}, with numbers taken from \citep{semanticfoam2026}.

\begin{figure}[t!]
    \centering
    \includegraphics[width=\linewidth, trim={0.5cm 5.5cm 0cm 3.5cm}, clip]{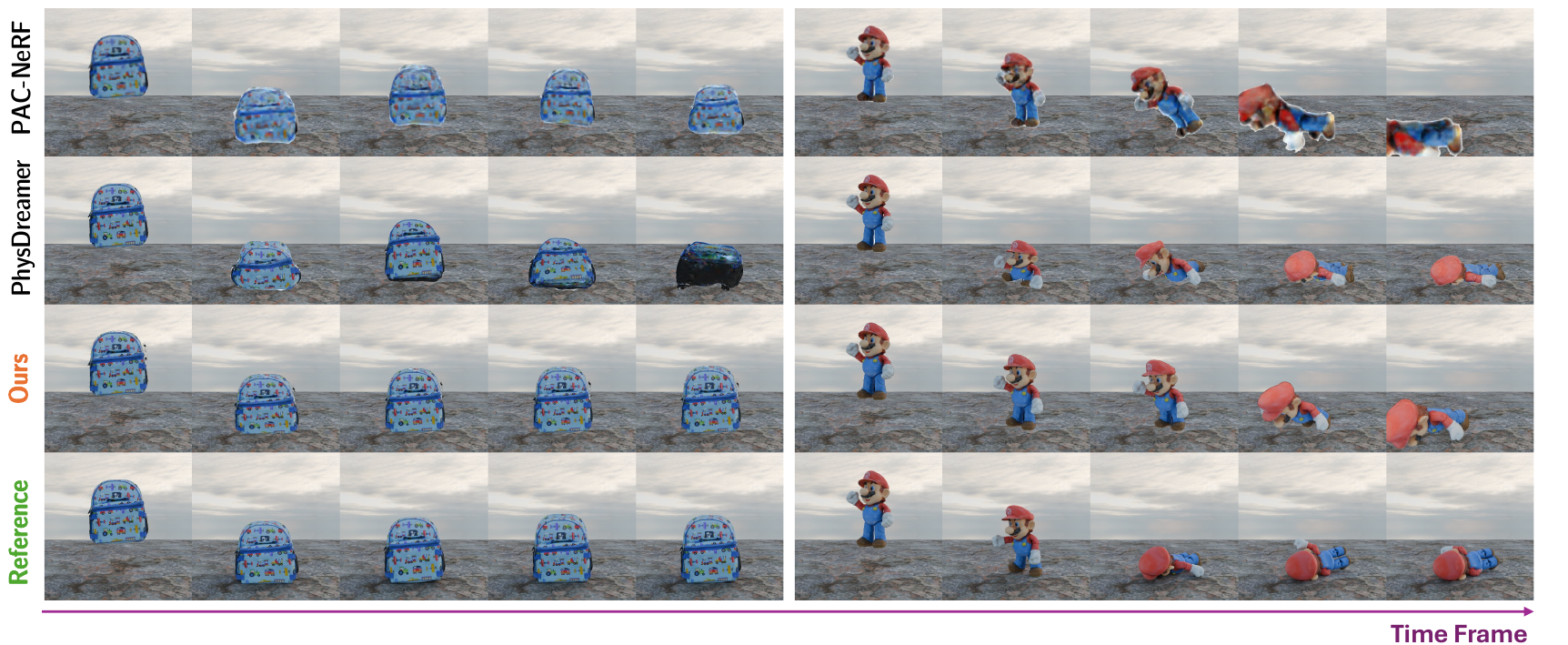}
\vspace{-15pt}
\caption{\textbf{Dynamic reconstruction under deformation and contact.}
\looseness=-1 PowerSim preserves the backpack's texture and Mario's shape as they fall and contact the ground. PAC-NeRF has blurred textures and distorted silhouettes, while PhysDreamer shows discrepancies in deformation and appearance. Reference frames are shown in the bottom row; time progresses from left to right.}
    \label{fig:qual_dynamic_rec}
\vspace{-5pt}
\end{figure}

\paragraph{Dynamic Reconstruction Evaluation.}
\looseness=-1 PowerSim achieves the highest mean PSNR and SSIM across the 12 objects (Tab.~\ref{tab:quan_dynamic_rec}). It leads in PSNR on 10 objects, reaching 25.73\,dB on average compared with 22.06\,dB for PAC-NeRF and 19.00\,dB for PhysDreamer. SSIM gains are smaller, with a mean of 0.930 versus 0.924 and 0.908, respectively. 
Fig.~\ref{fig:qual_dynamic_rec} highlights PowerSim's preservation of texture and object shape during motion, across both the deformable backpack and the nearly rigid Mario figure. Baselines exhibit texture blurring, shape distortion, or appearance artifacts.

\begin{table}[ht!]
\centering
\caption{\textbf{Dynamic reconstruction on 12 objects.}
\looseness=-1 PowerSim achieves the highest PSNR on 10 objects and SSIM on seven.
Bold and underline indicate the best and second-best results, respectively.}
\vspace{-5pt}
\setlength{\tabcolsep}{4pt}
\resizebox{\textwidth}{!}{%
\begin{tabular}{l|l|cccccccccccc|c}
\toprule
\textbf{Metrics} & \textbf{Method} & backpack & bell & blocks & bus & cream & elephant & grandpa & leather & lion & mario & sofa & turtle & Mean \\
\midrule
\multirow{3}{*}{PSNR $\uparrow$} & PAC-NeRF & \underline{19.37} & \textbf{25.00} & \underline{23.36} & \underline{20.72} & \underline{23.24} & \underline{22.27} & \underline{21.63} & \underline{20.85} & \underline{22.66} & \textbf{21.01} & \underline{22.49} & 22.19 & \underline{22.06} \\
& PhysDreamer & 18.93 & 19.54 & 19.79 & 18.82 & 19.83 & 17.14 & 16.91 & 18.27 & 18.28 & 17.56 & 19.58 & \underline{23.41} & 19.00 \\
& Ours & \textbf{23.69} & \underline{24.39} & \textbf{29.56} & \textbf{26.84} & \textbf{25.82} & \textbf{22.80} & \textbf{22.01} & \textbf{35.33} & \textbf{25.17} & \underline{20.31} & \textbf{24.72} & \textbf{28.11} & \textbf{25.73} \\
\midrule
\multirow{3}{*}{SSIM $\uparrow$} & PAC-NeRF & \underline{0.887} & \textbf{0.956} & \underline{0.940} & 0.908 & \underline{0.893} & \textbf{0.922} & \textbf{0.939} & 0.932 & \textbf{0.936} & \underline{0.921} & \textbf{0.926} & 0.923 & \underline{0.924} \\
& PhysDreamer & 0.856 & 0.940 & 0.917 & \underline{0.909} & 0.876 & 0.892 & \underline{0.922} & \underline{0.945} & 0.910 & 0.901 & 0.894 & \underline{0.939} & 0.908 \\
& Ours & \textbf{0.889} & \underline{0.944} & \textbf{0.953} & \textbf{0.938} & \textbf{0.923} & \underline{0.908} & 0.889 & \textbf{0.978} & \underline{0.930} & \textbf{0.941} & \underline{0.918} & \textbf{0.953} & \textbf{0.930} \\
\bottomrule
\end{tabular}%
}
\label{tab:quan_dynamic_rec}
\vspace{-10pt}
\end{table}

\paragraph{Material Estimation Evaluation.}
PowerSim achieves the lowest mean MAE in $\log(E)$, reducing the error from 0.60 for PhysDreamer to 0.44, a 27\% reduction (Tab.~\ref{tab:quan_material_estimate}). It achieves the lowest error on seven of the 12 objects, with particularly large gains on blocks, lion, and turtle. For Poisson's ratio, PowerSim matches PhysDreamer's mean MAE of 0.16 at the reported precision.

\begin{table}[ht]
\centering
\caption{\textbf{Material estimation on 12 objects.}
MAE in log Young's modulus and Poisson's ratio (lower is better).
Bold and underline indicate the best and second-best results, respectively.}
\vspace{-5pt}
\setlength{\tabcolsep}{4pt}
\resizebox{\textwidth}{!}{%
\begin{tabular}{l|l|cccccccccccc|c}
\toprule
\textbf{Metrics} & \textbf{Method} & backpack & bell & blocks & bus & cream & elephant & grandpa & leather & lion & mario & sofa & turtle & Mean \\
\midrule
\multirow{3}{*}{$\log(E)$} & PAC-NeRF & 3.28 & 1.08 & 4.02 & 3.30 & 3.22 & 3.05 & 2.99 & \underline{1.20} & 2.34 & 3.37 & \textbf{0.20} & 1.94 & 2.50 \\
& PhysDreamer & \underline{0.10} & \underline{1.07} & \underline{0.74} & \textbf{0.40} & \textbf{0.28} & \underline{0.54} & \textbf{0.43} & \textbf{0.33} & \underline{1.01} & \underline{1.18} & 0.68 & \underline{0.44} & \underline{0.60} \\
& \textbf{Ours} & \textbf{0.06} & \textbf{0.81} & \textbf{0.16} & \underline{0.60} & \underline{0.37} & \textbf{0.36} & \underline{0.47} & 1.21 & \textbf{0.20} & \textbf{0.60} & \underline{0.33} & \textbf{0.09} & \textbf{0.44} \\
\midrule
\multirow{3}{*}{$\nu$} & PAC-NeRF & \underline{0.21} & 0.23 & 0.33 & 0.16 & \textbf{0.12} & \underline{0.06} & 0.36 & \underline{0.26} & \underline{0.14} & \underline{0.33} & 0.30 & \textbf{0.01} & {0.21} \\
& PhysDreamer & 0.26 & \textbf{0.01} & \textbf{0.10} & \underline{0.14} & \underline{0.14} & \textbf{0.04} & \underline{0.33} & \textbf{0.11} & \textbf{0.06} & 0.44 & \textbf{0.05} & 0.29 & \textbf{0.16} \\
& \textbf{Ours} & \textbf{0.17} & \underline{0.18} & \underline{0.16} & \textbf{0.12} & 0.17 & 0.27 & \textbf{0.01} & \textbf{0.11} & 0.16 & \textbf{0.25} & \underline{0.09} & \underline{0.21} & \textbf{0.16} \\
\bottomrule
\end{tabular}%
}
\label{tab:quan_material_estimate}
\end{table}

\begin{figure}[t!]
    \centering  \includegraphics[width=\linewidth, trim={0.5cm 10cm 1cm 2.5cm}, clip]{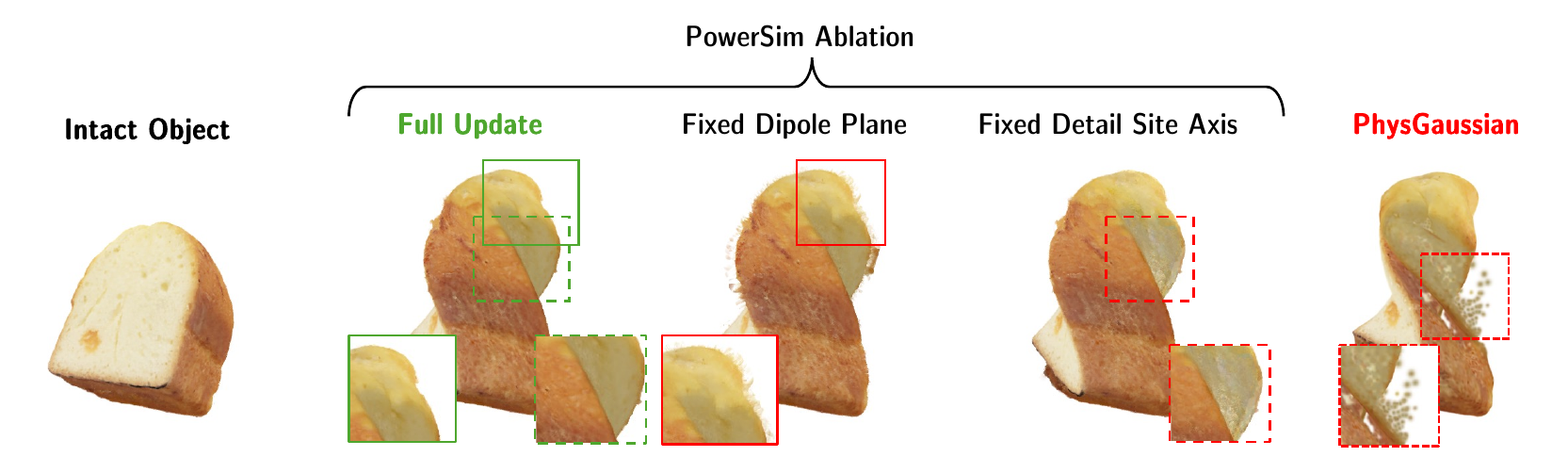}
\vspace{-20pt}
\caption{\textbf{Geometry and appearance updates under large deformation.}
Fixing dipole orientations produces ragged boundaries, while fixing appearance directions introduces color inconsistencies. The full PowerSim update preserves cleaner surface detail; PhysGaussian shows blurring and fragmentation in stretched regions. Insets highlight these differences. Best viewed on supplementary webpage.}
    \label{fig:ablate}
\vspace{-5pt}
\end{figure}

\paragraph{Ablation Studies.}
\label{subsec:ablation_studies}
\looseness=-1 We test role of geometry and appearance updates under large twisting and upward pulling (Fig.~\ref{fig:ablate}). Keeping dipole orientations fixed produces ragged boundaries and protruding surface fragments. Keeping appearance directions fixed introduces color inconsistencies as the surface rotates. Full update preserves cleaner boundaries and more consistent surface appearance throughout the motion. Under the same loading conditions, PhysGaussian shows blurred textures and fragmented stretched regions.

\paragraph{Additional Applications.}
With PowerSim, users can combine multiple objects into a single background scene and simulate all at once. Additionally, PowerSim supports both ray tracing and rasterization, opening up interesting application for physics simulation under complex secondary ray tracing effects. As demonstrated in Fig. \ref{fig:multi_physics}, our framework successfully handles multi-object interactions, capturing dynamic physics simulations and their reflections in a mirror environment.

\begin{figure}[t!]
    \centering  \includegraphics[width=\linewidth, trim={1cm 13cm 2cm 3cm}, clip]{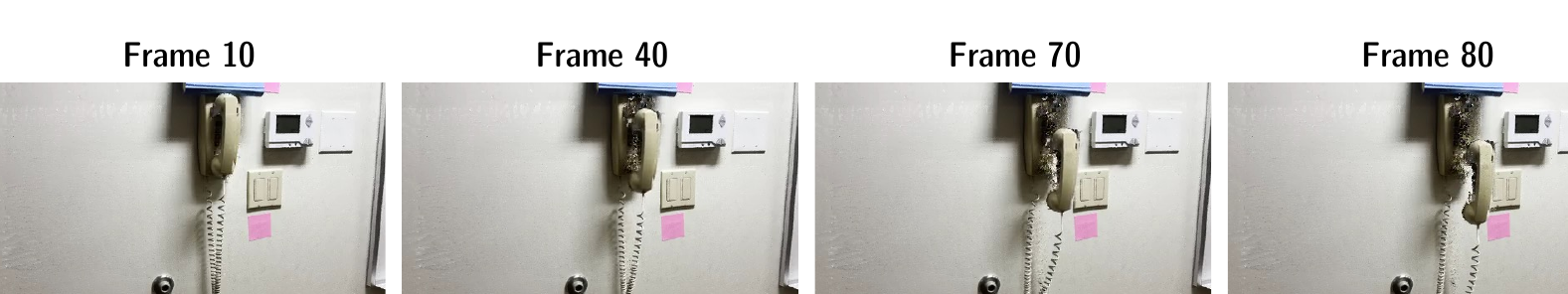}
\vspace{-5pt}
\caption{\textbf{Disocclusion artifacts.}
PowerSim can select and move the telephone handset, but its motion exposes poorly reconstructed regions, revealing gaps and visual artifacts.}
    \label{fig:limitation}
\vspace{-5pt}
\end{figure}

\section{Discussion}
\looseness=-1 PowerSim shows how a structured scene representation can support both physics simulation and rendering as objects deform. Updating primitive geometry, appearance and adjacency allows captured scenes to respond to new forces while retaining surface detail and supporting dynamic reflections. Differentiation through simulation and rendering also enables material estimation from deformation videos. Our coupling remains an approximation. Scalar radii capture isotropic scale changes, while dipole frames and appearance directions follow local rotation; individual primitives do not reproduce the full stretch and shear of the continuum. Although neighboring cells change shape as their sites move, this does not guarantee exact agreement between cell volumes and simulated material volumes. A richer deformation model could improve this correspondence. The results also depend on the captured geometry. Our selection method can isolate parts such as a telephone handset without optimization, but moving them may expose regions poorly observed during capture, revealing gaps and artifacts (Fig.~\ref{fig:limitation}). Generative priors could help complete these regions, which we leave to future work. Material estimation has a related ambiguity: different combinations of stiffness, initial velocity, and physical assumptions can explain similar image motion. The recovered parameters should therefore be interpreted under the assumed density, scale, and boundary conditions.

\subsection*{AI use statement}
In this work, we used generative AI tools for polishing and translating the manuscript text to English, producing initial drafts of some passages, condensing related literature, and configuring the software environments. The core research, including the research questions, the coupling between PowerFoam and MPM, the experimental design, and the implementation of PowerSim, was carried out by the authors without AI assistance; theoretical and proof-related uses do not apply to this paper. Every AI-assisted contribution was checked by the authors: generated text was revised by hand, environment setup scripts were validated by executing them and examining their outputs, and literature summaries were compared against the cited papers. We take responsibility for the final content of this work,
including text, claims or artifacts produced with the aid of generative AI.



\bibliography{iclr2027_conference}
\bibliographystyle{iclr2027_conference}
\newpage
\appendix
\section{Appendix}
\label{sec:appendix}
\subsection{Project Website}
We provide link to our project website at \url{https://power-sim.github.io/}.

\subsection{PowerFoam Geometry and Appearance Details}
\label{sec:powerfoam_details}

We provide the interpolation rules used by the PowerFoam
representation introduced in Section~\ref{sec:background}.
These rules are inherited from
PowerFoam~\citep{govindarajan2026powerfoam}.

\paragraph{Surface interpolation.}
The dipole plane separates the cell into a material side
opposite its normal $\mathbf{n}_i$, with density
$\boldsymbol{\sigma}_i \in \mathbb{R}_+$, and an empty side
with zero density.
The detail-site displacements bend this plane into a height
field. For an in-plane query
$\bar{\mathbf{x}}\in\mathbb{R}^2$, expressed in the same
radius-normalized coordinates as $\mathbf{s}_{i,k}$,
the weight of detail site $k$ is
\begin{equation}
w_{i,k}(\bar{\mathbf{x}})
=
\exp\!\left(
-\tau\|\bar{\mathbf{x}}-\mathbf{s}_{i,k}\|_2
\right),
\label{eq:pin_weight}
\end{equation}
where $\tau$ controls the sharpness of the interpolation.
The interpolated normal displacement is
\begin{equation}
h_i(\bar{\mathbf{x}})
=
\frac{
\sum_{k=1}^{K}w_{i,k}(\bar{\mathbf{x}})\,d_{i,k}
}{
\sum_{k=1}^{K}w_{i,k}(\bar{\mathbf{x}})
}.
\label{eq:height_blend}
\end{equation}
The corresponding surface point is
$\mathbf{p}_i
+\mathbf{r}_i(\bar{x}^{1}\mathbf{t}_i+\bar{x}^{2}\mathbf{b}_i)
+h_i(\bar{\mathbf{x}})\mathbf{n}_i$,
restricted to the bounded cell.

\paragraph{Directional appearance.}
Each detail site stores $L$ appearance directions
$\boldsymbol{\alpha}_{i,k}^{(l)}\in\mathbb{R}^3$
and corresponding colors
$\mathbf{c}_{i,k}^{(l)}\in\mathbb{R}^3$.
For a unit viewing direction $\mathbf{d}$, its color is
\begin{equation}
\begin{aligned}
\mathbf{c}_{i,k}(\mathbf{d})
&=
\frac{
\sum_{l=1}^{L}
w_{i,k}^{(l)}(\mathbf{d})\,\mathbf{c}_{i,k}^{(l)}
}{
\sum_{l=1}^{L}w_{i,k}^{(l)}(\mathbf{d})
},\\
w_{i,k}^{(l)}(\mathbf{d})
&=
\exp\!\left(
-\|\mathbf{d}-\hat{\boldsymbol{\alpha}}_{i,k}^{(l)}\|_2
\right),
\end{aligned}
\label{eq:sv_color}
\end{equation}
where
$\hat{\boldsymbol{\alpha}}_{i,k}^{(l)}
=
\boldsymbol{\alpha}_{i,k}^{(l)}
/
\|\boldsymbol{\alpha}_{i,k}^{(l)}\|_2$.
We then blend the detail-site colors using the spatial
weights from Eq.~\ref{eq:pin_weight}:
\begin{equation}
\mathbf{c}(\bar{\mathbf{x}},\mathbf{d})
=
\frac{
\sum_{k=1}^{K}
w_{i,k}(\bar{\mathbf{x}})\,\mathbf{c}_{i,k}(\mathbf{d})
}{
\sum_{k=1}^{K}w_{i,k}(\bar{\mathbf{x}})
}.
\label{eq:color_blend}
\end{equation}

\paragraph{Rest state and fixed attributes.}
For completeness, the reconstructed rest scene contains
\begin{equation}
\mathcal{S}_0
=
\left\{
\mathbf{p}_i^0,\mathbf{r}_i^0,\mathbf{q}_i^0,
\mathbf{s}_i,d_i,\boldsymbol{\sigma}_i,
\boldsymbol{\alpha}_i^0,\mathbf{c}_i
\right\}_{i=1}^{N},
\end{equation}
where $\mathbf{c}_i$ collects the colors
$\{\mathbf{c}_{i,k}^{(l)}\}_{k,l}$.
Simulation changes the positions, radii, frames, and appearance
directions. Detail-site coordinates, normal displacements,
densities, and color coefficients remain fixed.
In particular, Eq.~\ref{eq:pin_world_pos} scales the in-plane
offsets through $\mathbf{r}_i$, while the normal displacements
$d_{i,k}$ retain their original magnitudes.

\subsection{Material Point Method Simulation Loop}
\label{subsec:mpm_loop}
In this section, we provide additional background details on MPM simulation loop. We first define 
Lam\'e coefficients as:
\begin{equation} 
\mu = \frac{E}{2(1+\nu)}, \qquad \lambda = \frac{E\nu}{(1+\nu)(1-2\nu)}, \label{eq:lame} 
\end{equation} 
\paragraph{Particle-to-grid (P2G) transfer.} Mass and momentum are accumulated onto the grid with
the affine particle-in-cell (APIC) scheme \citep{jiang2015affine}, augmenting each particle's velocity with a
local affine velocity term $\mathbf{C}_p^t$:
\begin{align}
m_i^t &= \sum_p w_{ip}^t\, m_p, \\
(m\mathbf{v})_i^t &= \sum_p w_{ip}^t\, m_p
\left[ \mathbf{v}_p^t + \mathbf{C}_p^t \left(\mathbf{x}_i^t - \mathbf{x}_p^t\right) \right],
\end{align}
where $w_{ip}^t$ is the B-spline weight coupling particle $p$ to grid node $i$.

\paragraph{Grid update.} Each grid node velocity is integrated forward under the net nodal force $\mathbf{f}_i$, combining internal and external forces: 
\begin{equation} 
    \mathbf{v}_i^{t+1} = \mathbf{v}_i^t + \frac{\Delta t}{m_i^t}\, \mathbf{f}_i\!\left(\mathbf{x}_i^t; \boldsymbol\theta_p\right). 
\end{equation} The force follows from a hyperelastic energy $\Psi(\mathbf{F})$, and $\theta_p$ gathers the governing material properties, Young's modulus $E$, and Poisson's ratio $\nu$.
\paragraph{Grid-to-particle (G2P) transfer.} The updated nodal velocities are interpolated back to
the particles, whose positions are then advanced, the local affine velocity term is also updated correspondingly:
\begin{equation}
\mathbf{v}_p^{t+1} = \sum_i w_{ip}^t\, \mathbf{v}_i^{t+1}, \qquad
\mathbf{x}_p^{t+1} = \mathbf{x}_p^t + \Delta t\, \mathbf{v}_p^{t+1}, \qquad
\mathbf{C}_p^{t+1} = \frac{4}{(\Delta x)^2} \sum_{i} w_{ip}^t \mathbf{v}_i^{t+1} (\mathbf{x}_i - \mathbf{x}_p^t)^T
\end{equation}
\paragraph{Deformation gradient update.} Each particle's deformation gradient is updated from the
velocity gradient sampled off the grid:
\begin{equation}
\mathbf{F}_p^{t+1} = \left[ \mathbf{I}
+ \Delta t \sum_i \mathbf{v}_i^{t+1} \left(\nabla w_{ip}^t\right)^{\!\top} \right] \mathbf{F}_p^t.
\end{equation}

\subsection{Additional Implementation Details}
\label{sec:implementation details}
\paragraph{Material Field Optimization.} Following PhysDreamer \citep{zhang2024physdreamer}, we represent both velocity field and material field using smooth spatial fields: a triplane \citep{chan2022efficient}, queried at each primitive's rest position $\mathbf{p}_i^0$. For triplane representation, each field uses three $24\times24$ feature planes with 32 channels and a two-layer MLP decoder of width 64. The velocity field is zero-initialized so the object starts at rest. The material field, with weights $\phi$, outputs both components of $\boldsymbol{\theta}_p = (E_p,\nu_p)$. We optimize Young's in log-scale space around an initial guess $E$ and clamped to a plausible range $[1e2, 1e10]$, while Poisson's ratio is bounded to its physical interval $[0, 0.5]$. We
optimize with AdamW (weight decay $10^{-4}$) under a linear warmup over the first 10\% of
iterations followed by linear decay to zero, clipping the gradient norm to 1.0, with the
photometric loss of Eq.~\ref{eq:material_loss} at $\lambda = 0.2$. Stage 1 fits the velocity
field for 30 iterations on the first three frames at learning rate $10^{-2}$. Stage 2 freezes the velocity and fits the
material field on the full video for 20 iterations at $5\times10^{-3}$, initialized at
$\bar{E} = 10^{7}$~Pa.
\paragraph{Simulation Details.} We provide a list of constitutive models that we used for simulation in each scene in Tab.\ref{tab:model_settings}

\begin{table}[t]
\centering
\caption{\textbf{List of constitutive models used for simulation.}}
\label{tab:model_settings}
\begin{tabular}{lll}
\toprule
Scene & Figure & Constitutive Model \\
\midrule
Bonsai          & Fig.~\ref{fig:teaser} & Fixed corotated \\
Bread roll      & Fig.~\ref{fig:compare_with_physgauss} & Fixed corotated \\
Carnation       & Fig.~\ref{fig:material_estimate} & Fixed corotated \\
GSO benchmark   & Fig.~\ref{fig:qual_dynamic_rec} & Fixed corotated \\
Bread twist     & Fig.~\ref{fig:ablate} & Fixed corotated \\
Car             & Fig.~\ref{fig:multi_physics} & Fixed corotated \\
Can             & Fig.~\ref{fig:more_sim} & von Mises \\
Telephone       & Fig.~\ref{fig:more_sim} & Fixed corotated \\
Microphone      & Fig.~\ref{fig:more_sim} & Fixed corotated \\
Ficus           & Fig.~\ref{fig:more_sim} & Fixed corotated \\
Pillow          & Fig.~\ref{fig:more_sim} & Fixed corotated \\
Wolf            & Fig.~\ref{fig:more_sim} & Drucker--Prager \\
Paper Plane     & Fig.~\ref{fig:multi_physics} & Fixed corotated \\
\bottomrule
\end{tabular}
\end{table}

\begin{table*}[t]
    \centering
    \caption{Ablation studies on discounting factor $\beta$ with mIoU / mAcc per scene.}
    \label{tab:ablation_beta}
    \resizebox{\linewidth}{!}{%
    \begin{tabular}{lcccccc}
    \toprule
    $\beta$ & garden & bonsai & room & counter & kitchen & average \\
    \midrule
    0.1 & 0.912 / 0.948 & \textbf{0.837} / 0.881 & 0.604 / 0.953 & 0.758 / 0.912 & 0.855 / 0.947 & 0.793 / \textbf{0.928} \\
    0.2 & 0.916 / 0.948 & 0.833 / 0.877 & 0.608 / \textbf{0.955} & 0.758 / 0.912 & 0.855 / 0.947 & 0.794 / \textbf{0.928} \\
    0.4 & \textbf{0.922} / 0.948 & 0.826 / 0.868 & 0.615 / 0.953 & 0.759 / 0.912 & 0.858 / 0.947 & \textbf{0.796} / 0.926 \\
    0.5 & 0.920 / 0.945 & 0.822 / 0.863 & 0.619 / 0.953 & \textbf{0.759} / 0.912 & 0.859 / 0.947 & \textbf{0.796} / 0.924 \\
    0.6 & 0.920 / 0.944 & 0.819 / 0.860 & 0.622 / 0.952 & 0.759 / 0.912 & 0.859 / 0.947 & \textbf{0.796} / 0.923 \\
    0.8 & 0.917 / 0.940 & 0.814 / 0.853 & 0.625 / 0.948 & 0.759 / 0.912 & 0.859 / 0.947 & 0.795 / 0.920 \\
    1.0 & 0.916 / 0.938 & 0.810 / 0.848 & \textbf{0.630} / 0.939 & 0.759 / 0.912 & \textbf{0.861} / 0.947 & 0.795 / 0.917 \\
    \bottomrule
    \end{tabular}%
    }
\end{table*}

\vspace{-8pt}
\subsection{Additional Results}
\label{subsec:additional_results}
\begin{figure}[ht]
    \centering
    \includegraphics[width=\linewidth, trim={0.75cm 3cm 0cm 2cm}, clip]{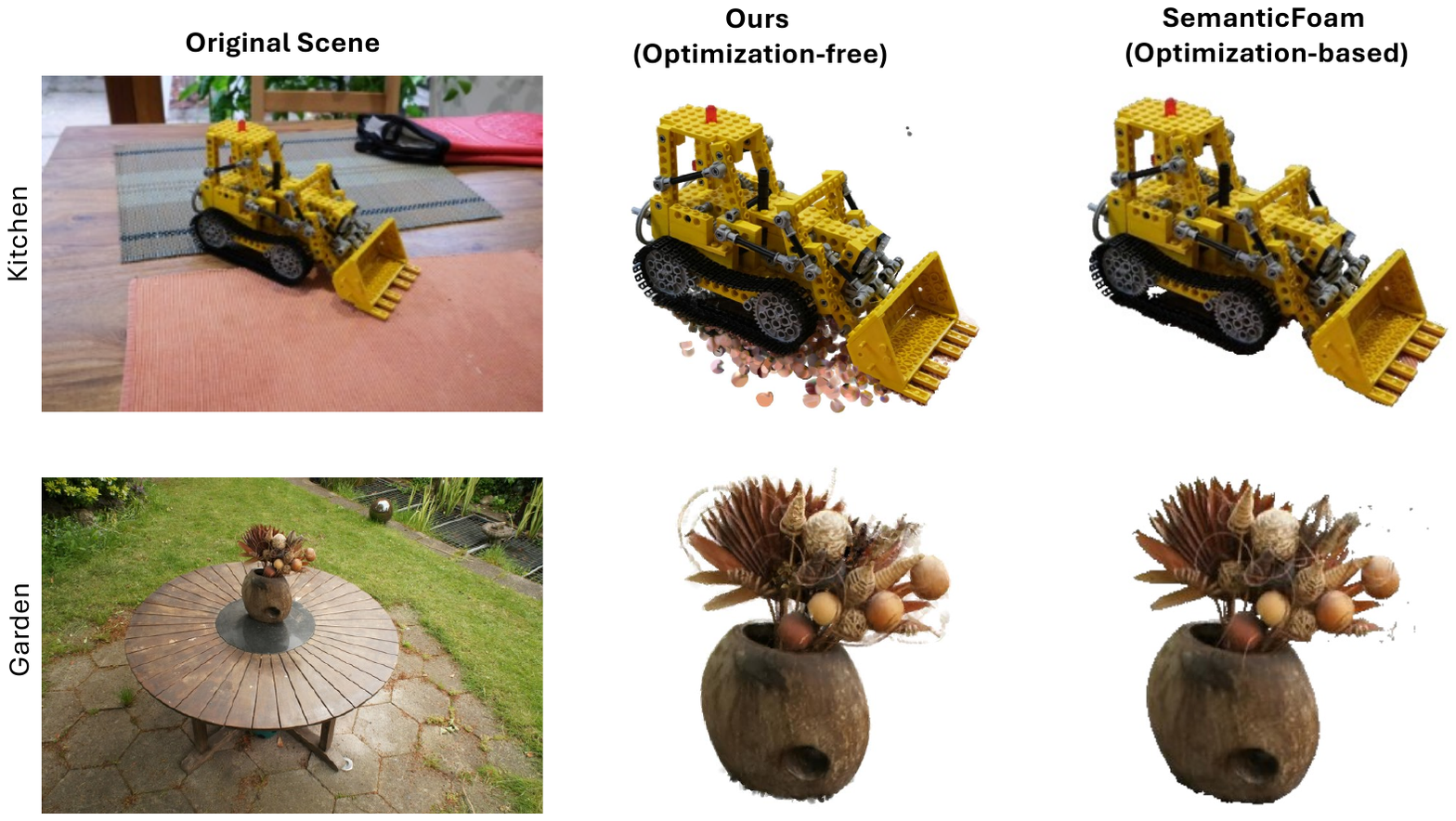}
    \caption{We provide qualitative results compared with SemanticFoam on Mip-NeRF 360 scenes}
    \label{fig:qual_foamedit}
\end{figure}

\paragraph{Additional Physics Simulation Results}
As shown in Fig.\ref{fig:more_sim}, we provide more simulation results spanning a range of objects, materials, and interactions. Our results are best viewed on our project website: \url{https://powersim.github.io/}.
\paragraph{3D Semantic Segmentation} To evaluate the performance of our proposed optimization-free dynamic primitive selection, we follow Semantic Foam\citep{semanticfoam2026} and evaluate on the 3D segmentation task using five Mip-NeRF 360 \citep{barron2022mip} for fair comparison with its released object masks, reporting mean Intersection over Union (mIoU) and mean Accuracy (mAcc). Tab.~\ref{tab:mipnerf360_seg} reports quantitative results. Without any optimization, our primitive selection is competitive with methods that train a dedicated semantic field. We additionally show objects extracted by our method alongside those of Semantic Foam's in Fig.\ref{fig:qual_foamedit}.
\paragraph{Ablation Studies on discounting factor.} Tab.\ref{tab:ablation_beta} varies the discounting factor $\beta$ from 0.1 to 1.0. We use $\beta = 0.5$
for all scenes.

\begin{table}[ht]
\centering
\caption{Per-scene segmentation result on Mip-NeRF~360. Baseline numbers are taken from the SemanticFoam paper. All baselines optimize a per-scene semantic field; ours requires \textbf{no optimization}.}
\resizebox{\textwidth}{!}{%
\begin{tabular}{lc|ccccc|c}
\toprule
\textbf{Method} & \textbf{Opt.-free} & \textbf{Garden} & \textbf{Bonsai} & \textbf{Room} & \textbf{Counter} & \textbf{Kitchen} & \textbf{Average} \\
 & & mIoU$\uparrow$ / mAcc$\uparrow$ & mIoU$\uparrow$ / mAcc$\uparrow$ & mIoU$\uparrow$ / mAcc$\uparrow$ & mIoU$\uparrow$ / mAcc$\uparrow$ & mIoU$\uparrow$ / mAcc$\uparrow$ & mIoU$\uparrow$ / mAcc$\uparrow$ \\
\midrule
LabelGS           & \xmark & 0.79 / \underline{0.95} & 0.70 / \underline{0.92} & \textbf{0.64} / \underline{0.93} & 0.55 / \textbf{0.93} & 0.82 / \textbf{0.96} & 0.70 / \textbf{0.94} \\
Gaussian Grouping & \xmark & 0.88 / 0.92 & 0.75 / 0.82 & 0.55 / 0.78 & 0.74 / \underline{0.92} & 0.83 / 0.92 & 0.75 / 0.87 \\
SemanticFoam      & \xmark & \textbf{0.94} / \textbf{0.96} & \textbf{0.90} / \textbf{0.94} & \underline{0.63} / \textbf{0.95} & \underline{0.75} / 0.91 & \textbf{0.90} / 0.94 & \textbf{0.82} / \textbf{0.94} \\
\midrule
Ours              & \cmark & \underline{0.92} / \underline{0.95} & \underline{0.82} / 0.86 & 0.62 / \textbf{0.95} & \textbf{0.76} / 0.91 & \underline{0.86} / \underline{0.95} & \underline{0.80} / \underline{0.92} \\
\bottomrule
\end{tabular}%
}
\label{tab:mipnerf360_seg}
\end{table}

\subsection{Assets Info}
\label{subsec:assets_info}
We provide more information about the assets that we use in the multi-object simulation. Other simulation scenes or objects are taken from PhysGaussian \cite{xie2023physgaussian} or PhysDreamer \cite{zhang2024physdreamer}.

\textbf{Car}: https://superspl.at/scene/bdb92f05

\textbf{Paper Plane}: https://sketchfab.com/3d-models/paper-plane-6e9201cd4b614d879741ea79524c57c6

\textbf{Can}: https://sketchfab.com/3d-models/french-coke-can-606cf0ae5cb44fa384bf2586982f7163

\end{document}